\documentclass[conference]{IEEEtran}
\pdfoutput=1
\usepackage[backend=biber,style=ieee]{biblatex}
\usepackage{hyperref}
\usepackage{amsmath,amssymb,amsfonts}
\usepackage{algorithmic}
\usepackage{graphicx}
\usepackage{textcomp}
\usepackage{xcolor}
\usepackage[inline]{enumitem}
\usepackage[utf8]{inputenc}
\usepackage[english]{babel}
\usepackage{cleveref}
\usepackage{booktabs}
\usepackage{makecell}
\usepackage{multirow}
\usepackage{todonotes}
\usepackage{siunitx}
\usepackage{xspace}
\usepackage{xargs}
\makeatletter
\let\storedcaption\@makecaption
\makeatother
\usepackage[font=footnotesize]{subcaption}
\makeatletter
\let\@makecaption\storedcaption
\makeatother

\usepackage{tikz}
\usetikzlibrary{positioning}
\usetikzlibrary{calc}
\usetikzlibrary{decorations.pathreplacing}

\usepackage{calc}
\usepackage{ifthen}
\newcommandx{\todonote}[3][1=]{\todo[linecolor=red,backgroundcolor=red!25,bordercolor=red,#1]{TODO (#2): #3}}
\newcommandx{\change}[2][1=]{\todo[linecolor=blue,backgroundcolor=blue!25,bordercolor=blue,#1]{#2}}
\newcommandx{\infonote}[3][1=]{\todo[linecolor=gray,backgroundcolor=gray!15,bordercolor=gray,#1]{#2: #3}}
\newcommandx{\final}[3][1=]{\todo[linecolor=green,backgroundcolor=green!15,bordercolor=green,#1]{#2: #3}}

\newcommandx{\tdzo}[2][1=]{\todonote[#1]{dzo}{#2}}
\newcommandx{\dzo}[2][1=]{\infonote[#1]{dzo}{#2}}
\newcommandx{\tada}[2][1=]{\todonote[#1]{ada}{#2}}
\newcommandx{\ada}[2][1=]{\infonote[#1]{ada}{#2}}
\newcommandx{\ideas}[2][1=]{\final[#1]{Idea}{#2}}
\newcommandx{\later}[2][1=]{\final[#1]{Later}{#2}}

\newcommand{\eg}{e.g.,\xspace}
\newcommand{\ie}{i.e.,\xspace}

\newcommand{\dota}{DotA\xspace}
\newcommand{\dotatwo}{Dota~2\xspace}
\newcommand{\dotatwos}{Dota~2\xspace}
\newcommand{\sdota}{Dota-350k\xspace}
\newcommand{\opendotas}{OpenDota\xspace}
\newcommand{\opendota}{OpenDota\xspace}
\newcommand{\pop}{Most Popular\xspace}
\newcommand{\pops}{POP\xspace}
\newcommand{\markov}{Markov\xspace}
\newcommand{\lr}{Logistic Regression\xspace}
\newcommand{\lrs}{LR\xspace}
\newcommand{\bertforrec}{BERT4Rec\xspace}
\newcommand{\sasrec}{SASRec\xspace}
\newcommand{\narm}{NARM\xspace}
\newcommand{\gru}{GRU\xspace}
\newcommand{\sir}{SIR\xspace}

\newcommand{\relu}{\ensuremath{\operatorname{relu}}}

\newcommand{\citeauthorwithcite}[1]{\citeauthor{#1}~\cite{#1}} 

\begin{document}
\def\review{0} 
\title{Sequential Item Recommendation in the MOBA Game \dotatwo}

\ifthenelse{\review = 1}{
\author{\IEEEauthorblockN{Anonymous Author}
\IEEEauthorblockA{\textit{Anonymous Chair} \\
\textit{Anonymous Institution}\\
Somewhere \\
anonymous@anonymous.author.com}
\and
\IEEEauthorblockN{Anonymous Author}
\IEEEauthorblockA{\textit{Anonymous Chair} \\
\textit{Anonymous Institution}\\
Somewhere \\
anonymous@anonymous.author.com}
\and
\IEEEauthorblockN{Anonymous Author}
\IEEEauthorblockA{\textit{Anonymous Chair} \\
\textit{Anonymous Institution}\\
Somewhere \\
anonymous@anonymous.author.com}
\and 
\IEEEauthorblockN{Anonymous Author}
\IEEEauthorblockA{\textit{Anonymous Chair} \\
\textit{Anonymous Institution}\\
Somewhere \\
anonymous@anonymous.author.com}
}}
{
\author{\IEEEauthorblockN{Alexander Dallmann}
\IEEEauthorblockA{\textit{Data Science Chair} \\
\textit{University of W\"urzburg}\\
W\"urzburg, Germany \\
dallmann@informatik.uni-wuerzburg.de}
\and
\IEEEauthorblockN{Johannes Kohlmann}
\IEEEauthorblockA{\textit{Data Science Chair}\\
\textit{University of W\"urzburg}\\
W\"urzburg, Germany \\
j.kohlmann@informatik.uni-wuerzburg.de}
\and
\IEEEauthorblockN{Daniel Zoller}
\IEEEauthorblockA{\textit{Data Science Chair} \\
\textit{University of W\"urzburg}\\
W\"urzburg, Germany \\
zoller@informatik.uni-wuerzburg.de}
\and
\IEEEauthorblockN{Andreas Hotho}
\IEEEauthorblockA{\textit{Data Science Chair} \\
\textit{University of W\"urzburg}\\
W\"urzburg, Germany \\
hotho@informatik.uni-wuerzburg.de}
}
}

\maketitle

\begin{abstract}

\textit{Multiplayer Online Battle Arena} (MOBA) games such as \dotatwo attract hundreds of thousands of players every year. 
Despite the large player base, it is still important to attract new players to prevent the community of a game from becoming inactive.
Entering MOBA games is, however, often demanding, requiring the player to learn numerous skills at once.
An important factor of success is buying the correct items which forms a complex task depending on various in-game factors such as already purchased items, the team composition, or available resources.
A recommendation system can support players by reducing the mental effort required to choose a suitable item, helping, \eg newer players or players returning to the game after a longer break, to focus on other aspects of the game.
Since \textit{Sequential Item Recommendation} (\sir) has proven to be effective in various domains (e.g. e-commerce, movie recommendation or playlist continuation), we explore the applicability of well-known SIR models in the context of purchase recommendations in \dotatwos.
To facilitate this research, we collect, analyze and publish \sdota, a new large dataset based on recent \dotatwo matches. 
We find that \sir models can be employed effectively for item recommendation in \dotatwo.
Our results show that models that consider the order of purchases are the most effective.
In contrast to other domains, we find RNN-based models to outperform the more recent Transformer-based architectures on \sdota.

\end{abstract}

\begin{IEEEkeywords}
Recommender systems, 
Recurrent neural networks,
Sequential recommendation,
E-Sports
\end{IEEEkeywords}

\section{Introduction}
\label{sec:introduction}

In the last decade, \textit{Multiplayer Online Battle Arena} (MOBA) games like Riot Games' \textit{League of Legends} or Valve`s \textit{\dotatwo} have continuously attracted hundreds of thousands of players every month\footnote{\url{https://steamcharts.com/app/570}}.
While most of the player base competes casually, there are also professional players and teams who participate in tournaments which yield massive prize pools.
The last \dotatwos world championships, \textit{The International 2019}\footnote{\url{https://www.dota2.com/international/overview}}, had a total prize pool of over \$34 million.
Furthermore, the live broadcast of this event was viewed by almost 2 million \dotatwos fans simultaneously\footnote{\url{https://www.statista.com/statistics/518201/dota2-championship-viewers/}}.
In order to foster an active community around MOBA games, it is important for publishers to attract a steady influx of new players to the game.
Learning to play MOBAs such as \dotatwo, however, is not a trivial task for inexperienced players since they are confronted with several new concepts at once.
For instance, a new player has to choose from a large pool of more than $100$ heros, navigate through battle and decide on which items he should spend his in-game currency to increase his effectiveness.
Making correct purchase decisions is essential for players to have significant impact on the outcome of a match.
However, deciding on the next purchase is a complex process that depends on various in-game factors such as already purchased items, allied and enemy heros, and the progress of the match.
Additionally, players have to choose from over 200 items in limited time due to the fast pace of \dotatwos matches. 
Therefore, a recommendation system can support players by reducing the mental effort required to choose a suitable item, helping, \eg newer players or players returning to the game after a longer break, to focus on other aspects of the game.

\begin{figure}
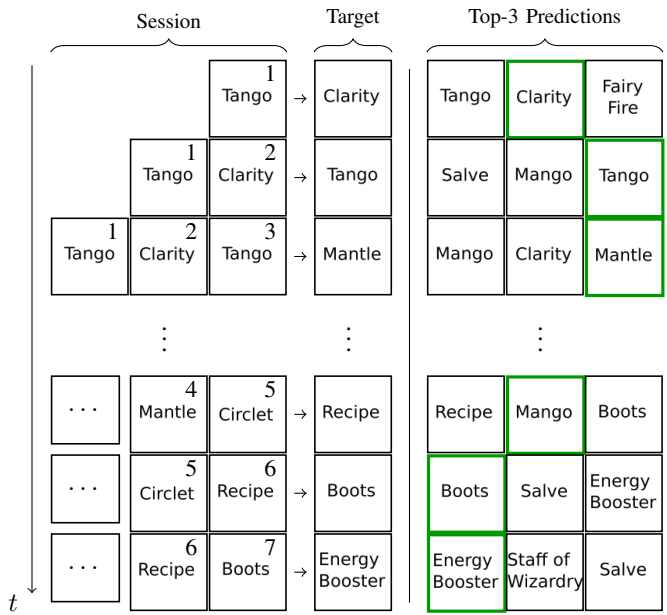

     \centering
     \begin{tikzpicture}

       \def\width{.9};
       \def\height{.9};
       \def\padding{.15};
       \def\u{cm};
       \def\w{2};
       \def\scale{0.32};
       \def\soffset{.5}
       \def\toffset{4}
       \def\poffset{5.5};
       \def\braceOffset{1mm};
       \def\mark{{1,2,2,0,1,0,0}};
       \def\dotrow{3};
       
       \tikzstyle{main}=[semithick, draw=black, minimum width = \width\u, minimum height=\height\u, inner sep=0pt,anchor = north west,align=center]
       \tikzstyle{pred}=[very thick, draw=black, minimum width = \width\u, minimum height=\height\u, inner sep=0pt,anchor = north west,align=center]
       
       
       \def\nb{6};
       \def\buys{{0,1,0,5,6,7,8,10}};

       \def\np{2};
       \def\pred{{{0,1,2},{3,4,0},{4,1,5},{0,0,0},{7,4,8},{8,3,9},{10,11,3}}};

       \foreach \i in {0,...,\nb} {
            \pgfmathsetmacro{\py}{0.1 + \i * (\height + \padding)};
            \ifthenelse{\i =\dotrow} {
                \pgfmathsetmacro{\px}{\soffset + \width + \padding};
                \node[main,draw=none] (dsession) at (\px, -\py) {$\vdots$};
                \node[main,draw=none] (dtarget) at (\toffset, -\py) {$\vdots$};
                \pgfmathsetmacro{\px}{\poffset +  \width + \padding};
                \node[main,draw=none] (dprediction) at (\px, -\py) {$\vdots$};
            }
            {
            \foreach \j in {0,...,\w} {
                \pgfmathsetmacro{\px}{\soffset + \j * \width +  \j * \padding};
                \pgfmathtruncatemacro{\itemidx}{\j + \i - \w};
                \pgfmathtruncatemacro{\wj}{\w - \j - 1};
                \pgfmathtruncatemacro{\il}{\i - (\w - \j) + 1};
                \pgfmathtruncatemacro{\nw}{\w + 1};
                \ifthenelse{\i < \nw}{
                
                \ifthenelse{\wj < \i}{
                    \pgfmathsetmacro{\item}{\buys[\itemidx]};
                    \node[main,label={[black, anchor=north west,xshift=1mm,yshift=.5mm]\small\il}] (s\i-\j) at (\px,-\py) {\includegraphics[scale=\scale]{assets/items/\item}};
                    }
                    {
                        \node[main,draw=none] (s\i-\j) at (\px,-\py) {};
                    }
                }
                {
                    \ifthenelse{\j = 0} {
                         \node[main] (s\i-\j) at (\px,-\py) {$\dots$};
                    }
                    {
                        \pgfmathsetmacro{\item}{\buys[\itemidx]};
                        \node[main,label={[black, anchor=north west,xshift=1mm,yshift=.5mm]\small\il}] (s\i-\j) at (\px,-\py) {\includegraphics[scale=\scale]{assets/items/\item}};
                    }
                }
                }
            
            \pgfmathsetmacro{\target}{\buys[\i + 1]};
            \node[main] (t\i) at (\toffset,-\py) {\includegraphics[scale=\scale]{assets/items/\target}};
            \draw[->] ([xshift=1mm]s\i-\w.east) to ([xshift=-1mm]t\i.west);
            
            \pgfmathsetmacro{\shouldMark}{\mark[\i]};
            \foreach \j in {0,...,\np} {
                \pgfmathsetmacro{\px}{\poffset + \j * (\width + \padding)};
                \pgfmathsetmacro{\item}{\pred[\i][\j]};
                \ifthenelse{\shouldMark = \j}
                {\node[pred,black!40!green] (p\i-\j) at (\px,-\py) {\includegraphics[scale=\scale]{assets/items/\item}}}
                {\node[main] (p\i-\j) at (\px,-\py) {\includegraphics[scale=\scale]{assets/items/\item}}};
            }
            }
        }
        
        \node (st) at ($(t0.north east)!0.5!(p0-0.north west)$) {};
        \node (se) at ($(t\nb.south east)!0.5!(p\nb-0.south west)$) {};
        \draw[-] (st) to (se);
        
       \draw [decorate,decoration={brace,amplitude=5pt}]
        ([yshift=\braceOffset]s0-0.north west) -- ([yshift=\braceOffset]s0-\w.north east) node [midway, above, yshift=.2cm] {\footnotesize Session};
       \draw [decorate,decoration={brace,amplitude=5pt}]
        ([yshift=\braceOffset]t0.north west) -- ([yshift=\braceOffset]t0.north east) node [midway, above, yshift=.2cm] {\footnotesize Target};
       \draw [decorate,decoration={brace,amplitude=5pt}]
        ([yshift=\braceOffset]p0-0.north west) -- ([yshift=\braceOffset]p0-\np.north east) node [midway, above, yshift=.2cm] {\footnotesize Top-3 Predictions};
        
        \node (ts) at ($(0,0)!0.5!(s0-0.north west)$) {};
        \node (ter) at ($(s\nb-0.south west)$) {};
        \node (tel) at (ter -| {{(0,0)}}) {$t$};
        \node (te) at ($(tel)!0.5!(ter)$) {};
        \draw[->] (ts) to (te);
        
     \end{tikzpicture}
     \caption{Overview of the item recommendation process in the context of \dotatwos. The first column indicates the items currently present in the player's inventory with the index in the upper-right corner indicating when the item was bought. The second column shows the next actual purchase while the last column depicts the top-3 predictions of the recommendation system. Correct predictions are marked green. Item respresentations are inspired by Dota 2\cite{valve}.}
     \label{fig:buy_and_predictions}
 \end{figure}

Since the purchase of items in \dotatwos is an inherently sequential process, that is, items are bought one after another with the next purchase depending on the items already present in the players inventory (see \Cref{fig:buy_and_predictions}), we propose the use of \textit{Sequential Item Recommendation} (\sir) as a first step towards item recommendation in \dotatwos matches.
\sir has been applied successfully in various domains such as e-commerce, playlist continuation and movie recommendation \cite{chen2018recsys, kang2018selfattentive, sun2019bert4rec,tan2016improved,li2017neural}.
It focuses on recommending an item to a user based on previous interactions they had with other items \cite{fang2020deep}.
Recently, neural models for \sir have shown superior performance on a variety of datasets \cite{tan2016improved,li2017neural, kang2018selfattentive, sun2019bert4rec} and are now widely used.
In this work, we introduce a large new dataset (\sdota) that contains in-game item purchases and interesting features, \eg purchase timings, the played hero and a comparatively small item set, and compare it to an older and smaller dataset (\opendota).
We show that neural sequential models outperform statistical and non-sequential baselines previously used in this setting and that they yield improved performance compared to a recently proposed model based on logistic regression \cite{looi2019recommendation} on both datasets.
Our results show that RNN-based \sir models yield good results, outperforming transformer-based models and achieving high scores in terms of Rec@3 = $0.7363$ and NDCG@3 = $0.6307$ on \sdota. 

Our contributions in this paper are:
\begin{enumerate}
    \item We introduce \sdota\footnote{\url{https://professor-x.de/papers/dota-350k}}, a large dataset comprised of purchase sessions from recent \dotatwo matches and compare it to a smaller and older dataset from \opendota.
    \item We provide an analysis of the dataset in the context of sequential item recommendation.
    \item We compare the performance of state-of-the-art \sir models and find that, in contrast to datasets from other domains, recurrent models outperform transformer models.
\end{enumerate}
The rest of the paper is structured as follows: We briefly explain the core mechanics of \dotatwos in \Cref{sec:background}. Afterwards,
we give a short overview of related work in \Cref{sec:related_work}.
Then, we give a detailed description of the dataset and its properties in \Cref{sec:dataset}.
This is followed by a description of our experiments and their results including a discussion of our findings in \Cref{sec:experiments}.
Finally, we provide a compilation of our results and possible future work in \Cref{sec:conclusion}.

\section{Background: \dotatwo}
\label{sec:background}
\textit{\dotatwo} is a (MOBA) game released by \textit{Valve} \cite{valve}.
The game is played in matches between two teams of five players on a static map.
The goal of the game is to destroy the \textit{ancient}, a large structure in the opposing team base.
To achieve this goal, each player controls a character, that is, a \textit{hero} with a unique set of abilities and a unique play style.
During the match the players earn the in-game currency \textit{gold} by defeating the enemy's heroes in combat or destroying enemy buildings.
The earned gold can be used to purchase \textit{items}, which either improve the basic hero properties, for example, strength, intelligence, and agility directly or unlock additional active or passive skills of a hero.
The space for these items is limited to nine in \dotatwo for a hero.\footnote{At most six of those items can be worn actively, while the other three are only stored and do not affect the hero.}
Some items can be combined to yield a new, more powerful item.

While most items affect the hero permanently, \dotatwo also includes \textit{consumable} items.
These items can be activated, consuming the item and granting an effect to the hero that usually wears off after a short period of time. Besides affecting the consuming hero, some consumables have other special effects. Different types of \textit{wards} can, for instance, be placed on the map to reveal a small area around them to the player and their allies for up to seven minutes. Since information about enemy movement is critical to fight them effectively, \textit{wards} are purchased frequently throughout every match.

In contrast to most offline games, \dotatwos constantly receives updates to its in-game content in the form of \textit{patches}. While \textit{patches} sometimes introduce new items or heros, they are mostly used to re-balance the strength of already existing ones. These changes heavily influence the in-game \textit{meta}, i.e. which combinations of heros and items are popular among players.

\section{Related Work}
\label{sec:related_work}
In this section we give a brief overview of related research into \dotatwos as well as Sequential Item Recommendation.

\subsection{\dotatwo Research Interests}
Over the last decade, research interest into MOBA games such as \dotatwos has grown continuously. Since the main goal in a \dotatwos match is defeating the opposing team to win the match, most research efforts explore the prediction of the winning team based on several pre- and in-game factors \cite{Akhmedov2021MachineLM,Conley2013HowDH,kinkade2015dota,semenov2016dota2,Wang2016PredictingMO,yang2016realtime}. Early work in this context mostly focused on using draft data of both teams to train logistic regression (LR) and $k$-Nearest-Neighbor classifiers for win probability prediction \cite{Conley2013HowDH,Wang2016PredictingMO}. Although draft data allowed for accuracy's up to 70\%, the experiments showed that further information was needed to achieve more accurate predictions. Therefore, several additional features like hero and match statistics, and team synergies, were introduced simultaneously with more complex models \cite{kinkade2015dota,Wang2016PredictingMO,yang2016realtime,zhang2020lineup}. Besides outcome prediction, there have been occasional explorations of other aspects of \dotatwos such as encounter detection, spatio-temporal movement analysis and role classification \cite{harrison2014patterns,rioult2014mining,drachen2014spatiotemporal,eggert2015classification,demediuk2019classification}. Item recommendation in \dotatwos was, however, only recently first addressed by \citeauthorwithcite{looi2019recommendation}. They used association rules and \lr (\lrs) to recommend items to the player based on their purchase history.

\subsection{Sequential Item Recommendation}
Recently, recommendation algorithms have become prevalent in every aspect of life, for instance, when deciding on a movie to watch, a restaurant to go to, or an item to buy.
Since user behavior heavily depends on previous interactions, \sir considers the time-ordered interaction history of each user to provide recommendations~\cite{fang2020deep,wang2019recommender,tan2016improved,li2017neural,tang2018personalized,kang2018selfattentive,sun2019bert4rec}.
Due to the wide applicability of \sir and the recent advances in deep learning, many neural \sir systems have been proposed.
While these models previously relied on recurrent or convolutional neural networks (\eg \cite{tan2016improved,li2017neural,tang2018personalized}), the focus recently shifted towards self-attentative models (\eg \cite{kang2018selfattentive,sun2019bert4rec}) with the Transformer architecture \cite{vaswani2017attention} gaining popularity.
\sir models are predominantly evaluated on datasets from the e-commerce (\eg \textit{Yoochoose, Amazon Beauty, Steam}) or movie rating (\eg \textit{Movielens-1M, Movielens-20M}) domain \cite{tan2016improved,sun2019bert4rec,kang2018selfattentive,tang2018personalized}.

Despite \dotatwos research and \sir recently experiencing gains in popularity, there has not been any research into applying modern, neural SIR systems to item recommendation in \dotatwos matches.
This might partly be caused by the fact that there was no large-scale, easily available and relatively recent dataset of \dotatwos purchase events.
\section{Problem Setting}
\label{sec:setting}
Based on the history of a user's item interactions, the aim of a \sir model is to predict the next interaction of the user.
Translated into the context of item buy predictions in \dotatwo, the model predicts the next item a player is going to buy based on past purchases during a match.

More formally, let $M$ be a set of matches and $I = \{i_1, i_2, \ldots, i_{|I|}\}$ the set of items that can be bought during the match, $U =\{u_1, u_2, \ldots, u_{|U|}\}$ the set of unique players and $S$ the set of all possible sessions (i.e., the purchases a player makes during a match) in the dataset.
For each participating player, we extract a session $s^u_{m} = [s^u_{m,1}, s^u_{m,2}, \ldots, s^u_{m,l_s}] \in S$ where $m \in M$ denotes the match, $u \in U$ denotes the player participating in the match, $s^u_{m, j} \in I$ is the $j$-th item bought during the match and $l_s$ is the total number of items bought by the player in this session.
The \sir model can then be formulated as $K: S \to I \times \mathbb{R}^+_0$ where $K(s) = (i, p)$ indicates that item $i$ was assigned score $p$ by the model $K$ based on the session $s$.
\section{Dataset}
\label{sec:dataset}
In this section, we describe the already existing OpenDota dataset and our new \sdota dataset.
We detail how we obtained and preprocessed the data and present key statistics of the datasets that are relevant for training \sir models.
\subsection{\sdota Dataset}
For this paper, we collected a dataset of \dotatwo matches played on the 16th of April 2020.\footnote{\dotatwo Patch Version: 7.25c}
\paragraph{Collection}
We use the official API\footnote{\url{http://api.steampowered.com/IDOTA2Match\_570/GetMatchHistoryBySequenceNum/v1}} to collect summary data for all publicly available matches on this day.
Since the game rules vary between modes, we restrict our dataset to matches from the most played game mode \textit{Ranked All Pick} which accounts for $54.13\%$ of all matches played on the 16th of April 2020.
Since \textit{Ranked} matches are played for \textit{Ranking Points} (MMR), we can assume that players will play more seriously, reducing noise in the dataset.

Unfortunately, players sometimes abandon a match, by either stopping to play, actively disconnecting from the game, or losing their internet connection for too long.
This is a common problem in \dotatwo and although penalties for such behaviour exist, it still affects \num{17.70}{\%} of all \textit{Ranked All Pick} matches in our dataset.
Since the absence of a teammate changes the game dynamics, we do not add these matches to the dataset.

We proceed to download the game replay file for all selected matches.
The replay file contains all information necessary to reconstruct the entire match.
Using the \textit{Clarity} parser,\footnote{\url{https://github.com/skadistats/clarity}}  sessions for every user in the game are extracted by tracking the item purchases during the match.
Additionally, the dataset contains the played \textit{hero} and \textit{team affiliation} for each session.
It is common in the \sir setting to remove sessions with only one item ($l_s < 2$).
Instead of only removing the affected session, we discard the entire match since the occurrence of an abnormally short session causes the remaining players to adapt to the underperforming teammate, by adjusting their itemization.
Additionally, we removed $24$ matches where an item that is not purchasable appeared in the purchase history of at least one player. For a very small fraction of games, less than $10$ sessions could be extracted during processing of the replay file. Since these are otherwise normal ranked matches, we decided to keep those matches. Afterwards, we sort all remaining matches by duration and remove the top and bottom 2.5\% quantile. We take this step to exclude unusually short or long matches since these are often caused by abnormal game events such as players giving up or actively sabotaging their allies.

\paragraph{Basic Statistics and Split}
The basic properties of the dataset are listed in \Cref{tab:dataset_statistics}. 
\begin{table}
    \centering
    \caption{Overview of the basic properties and the individual splits of both datsets.}
    \begin{subtable}{\linewidth}
        \centering
        \begin{tabular}{lrr|rrr}
            \toprule
            & Raw &   Processed&   Training &   Validation &      Test \\
            \midrule
            Fraction                 & -       & 100\%     & 94\%      & 1\%   & 5\%   \\
            \# Matches               & 590799  & 348642    & 327723    & 3486  & 17433 \\
            \# Sessions              & 5907396 & 3486358   & 3277173   & 34859 & 174326\\
            \# Items                 & 247     & 212       & 212       & 212   & 212   \\  
            \# Heros                 & 119     & 119       & 119       & 119   & 119   \\
            $\mu(l_s)$               & 45.37   & 43.27     & 43.30     & 43.56 & 42.66 \\
            $\sigma(l_s)$            & 19.07   & 13.89     & 13.89     & 14.15 & 13.80 \\
            \bottomrule
        \end{tabular}
        \caption{\sdota}
        \label{tab:dataset_statistics}
    \end{subtable}

    \begin{subtable}{\linewidth}
        \centering
        \begin{tabular}{lrr|rrr}
            \toprule
            & Raw &   Processed &   Training &   Validation &     Test \\
            \midrule
            Fraction                   & -     &  100\%    & 90\%      & 5\%       & 5\%       \\
            \# Matches                 & 50000 & 39921     & 35928     & 1996      & 1997      \\
            \# Sessions                & 499943& 399185    & 359256    & 19959     & 19970     \\
            \# Items                   & 146   & 141       & 141       & 141       & 141       \\        
            \# Heros                   & 110   & 110       & 110       & 110       & 110       \\
            $\mu(l_s)$                 & 36.39 & 35.20     & 35.21     & 35.11     & 35.14     \\
            $\sigma(l_s)$              & 11.54 & 9.58      & 9.58      & 9.54      & 9.52      \\
            \bottomrule
        \end{tabular}
        \caption{\opendota}
        \label{tab:opendota_statistics}
    \end{subtable}
    \vspace{-2em}
\end{table}

To evaluate the models close to a real-world application, where the recommender can only be trained on past games,
we order the matches in the final dataset by start time and split the dataset by selecting the first 94\% as training, the following 1\% as validation and the final 5\% as the test set.
Note that the dataset is split by matches as opposed to sessions to avoid information leaking across different sessions.
Instead of using equally sized validation and test splits, we build a relatively small validation set and add the remaining data to the training set.
The reasoning behind this step is two-fold: First of all, a small validation set increases the amount of data available for training while keeping the test set large enough to give reliable evaluation results.
Secondly, it decreases the computational demands of the validation step dramatically allowing for faster training and extensive hyperparameter studies.
To ensure that the validation data is still representative, we increased the split size incrementally until it contained all items and heros.
Furthermore, we ensured that key session characteristics such as average session length and standard deviation match those of the whole dataset closely.
Finally, we computed the \textit{Kendall-Tau} rank correlation coefficient\cite{kendall1938tau} for the items ranked by purchase frequency and heros ranked by pick-rate between the validation set and the whole dataset.
We obtained correlations of $98.30\%$ and $92.25\%$ respectively, indicating that 1\% of the data is sufficient to accurately represent the whole dataset.

\begin{figure}
    \centering
    \includegraphics[scale=0.5]{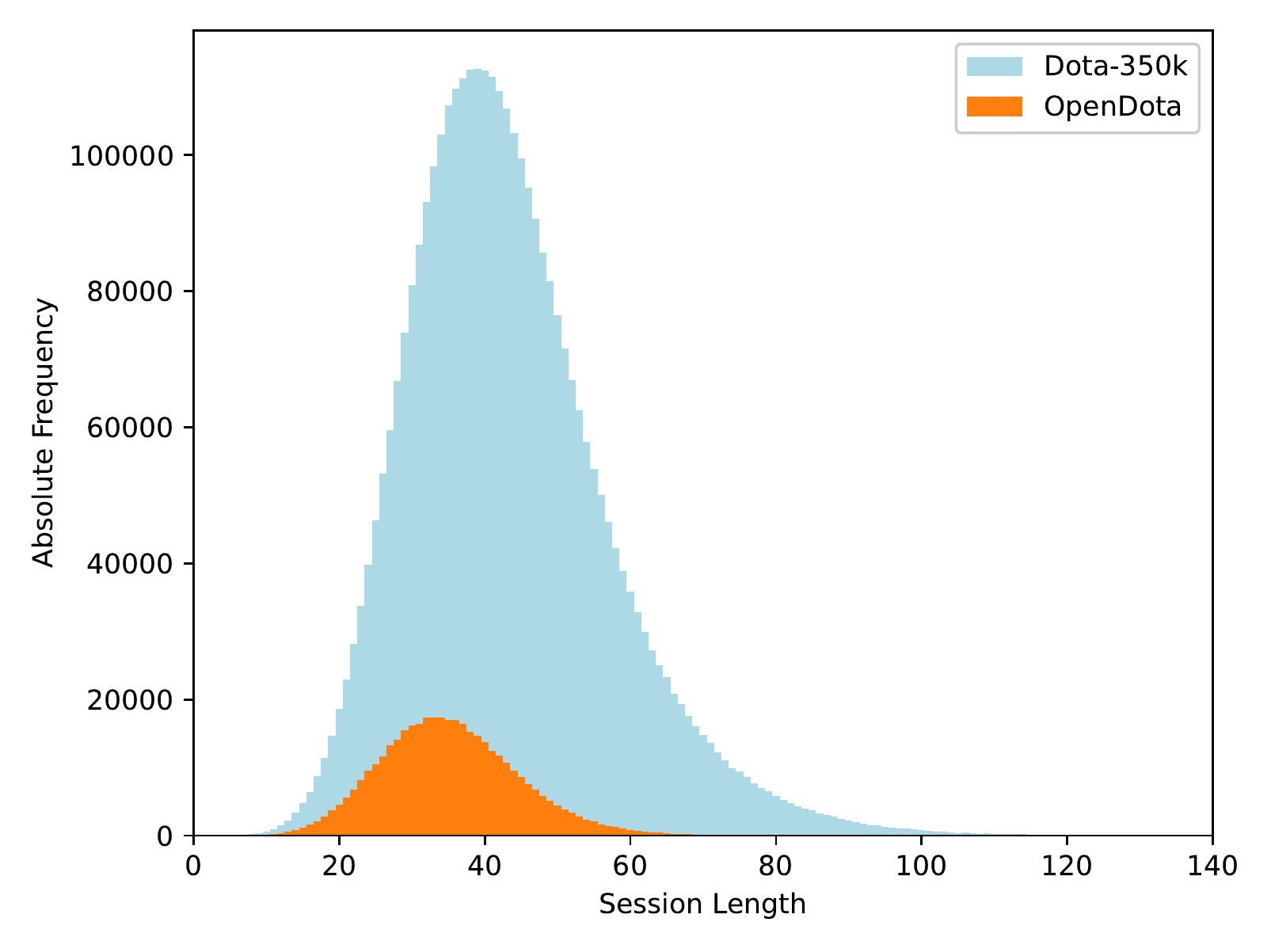}
    \caption{Distribution of the number of items bought (\textit{session length}) over all matches in both datasets.}
    \label{fig:session_length}
\end{figure}

In contrast to other purchase data, \sdota has a relatively small item space of 212 unique items.
Furthermore, the item purchase frequency distribution in \sdota does not feature the usual long-tail, instead even items with a comparatively low frequency are bought several thousand times (see \Cref{fig:item_and_hero_frequencies}).
Additionally, the session length frequencies follow an approximately normal distribution with mean $\mu = 43.27$ and standard deviation $\sigma = 13.89$ (see \Cref{fig:session_length}).
The distribution has a slight tendency towards longer sessions, which intuitively makes sense:
The number of items a player can buy increases with the duration of the match. However, the limited match duration hinders the occurrence of very long sessions. 
The likelihood of very short sessions occurring is low due to players spending their initial gold at the start of the game.
Also, the removal of games with a short duration ensures that the remaining sessions always include several purchases.
The most frequently bought items in \sdota are \textit{consumables} (see \Cref{fig:item_and_hero_frequencies}).
Since consumables are relatively cheap, often bought several times per session and include key items like \textit{wards}, it is reasonable for them to lead the purchase ranking.
The unique properties of consumable items also influence their purchase times.
Due to their low cost, they are often bought at the start of the match (see \Cref{fig:ward_time_dist}).
In contrast, more expensive items like \textit{Ethereal Blade} are bought later during the match (see \Cref{fig:blade_time_dist}).
This is intuitively clear, since players need to accumulate enough gold before they are able to purchase expensive items.

\subsection{\opendota Dataset}
\begin{figure}
    \centering
    \includegraphics[scale=0.5]{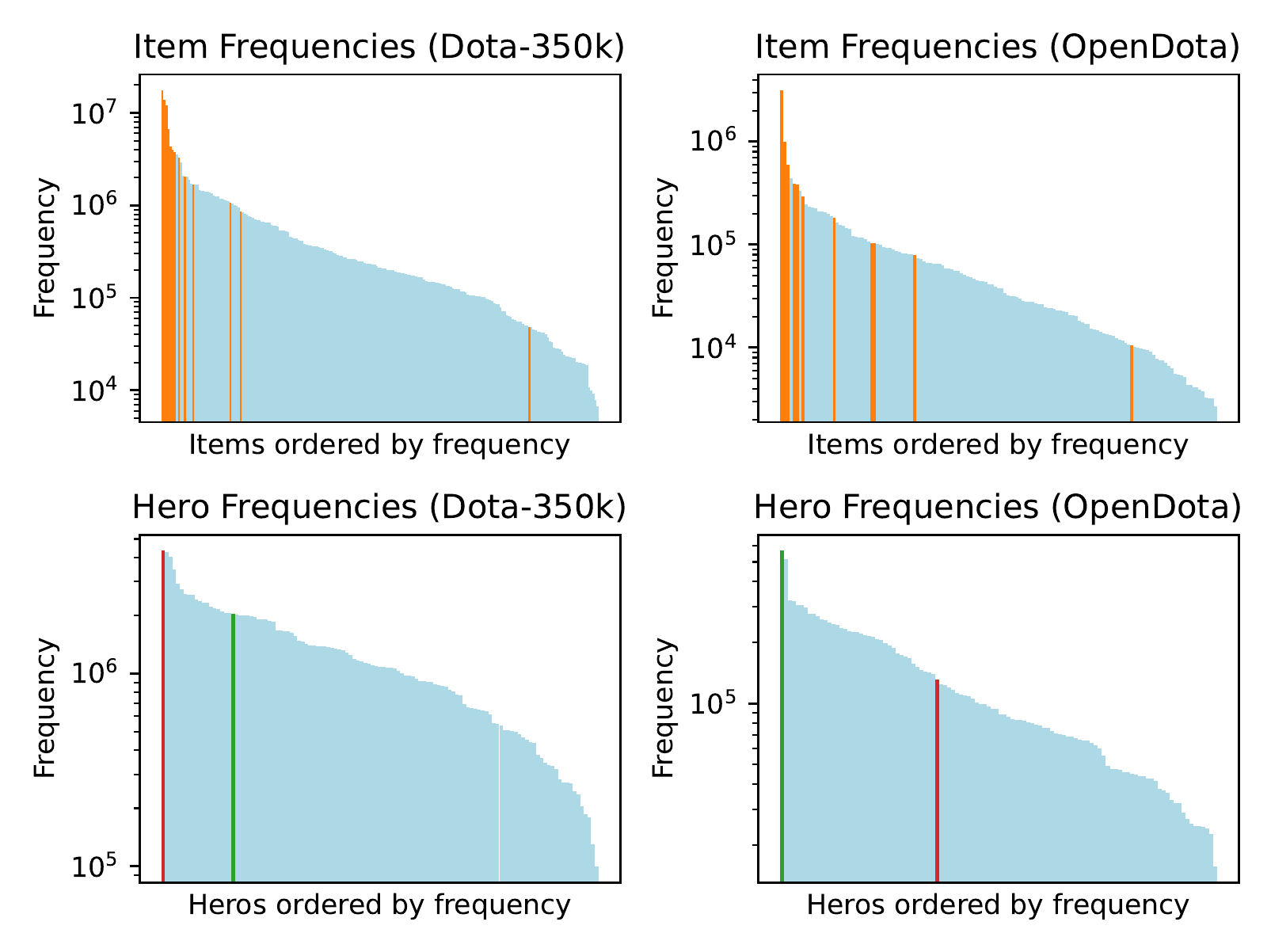}
    \caption{Side-By-Side comparison of the buy frequencies of each item and the pick frequency of each hero. Consumable items are marked orange. The heros \textit{Ogre Magi} (red) and \textit{Windranger} (green) are marked in the bottom row to visualize the \textit{meta} changes between both datsets.}
    \label{fig:item_and_hero_frequencies}
\end{figure}

\begin{figure} 
    \centering
  \subfloat[Item Sentry Ward\label{fig:ward_time_dist}]{%
       \includegraphics[scale=0.5]{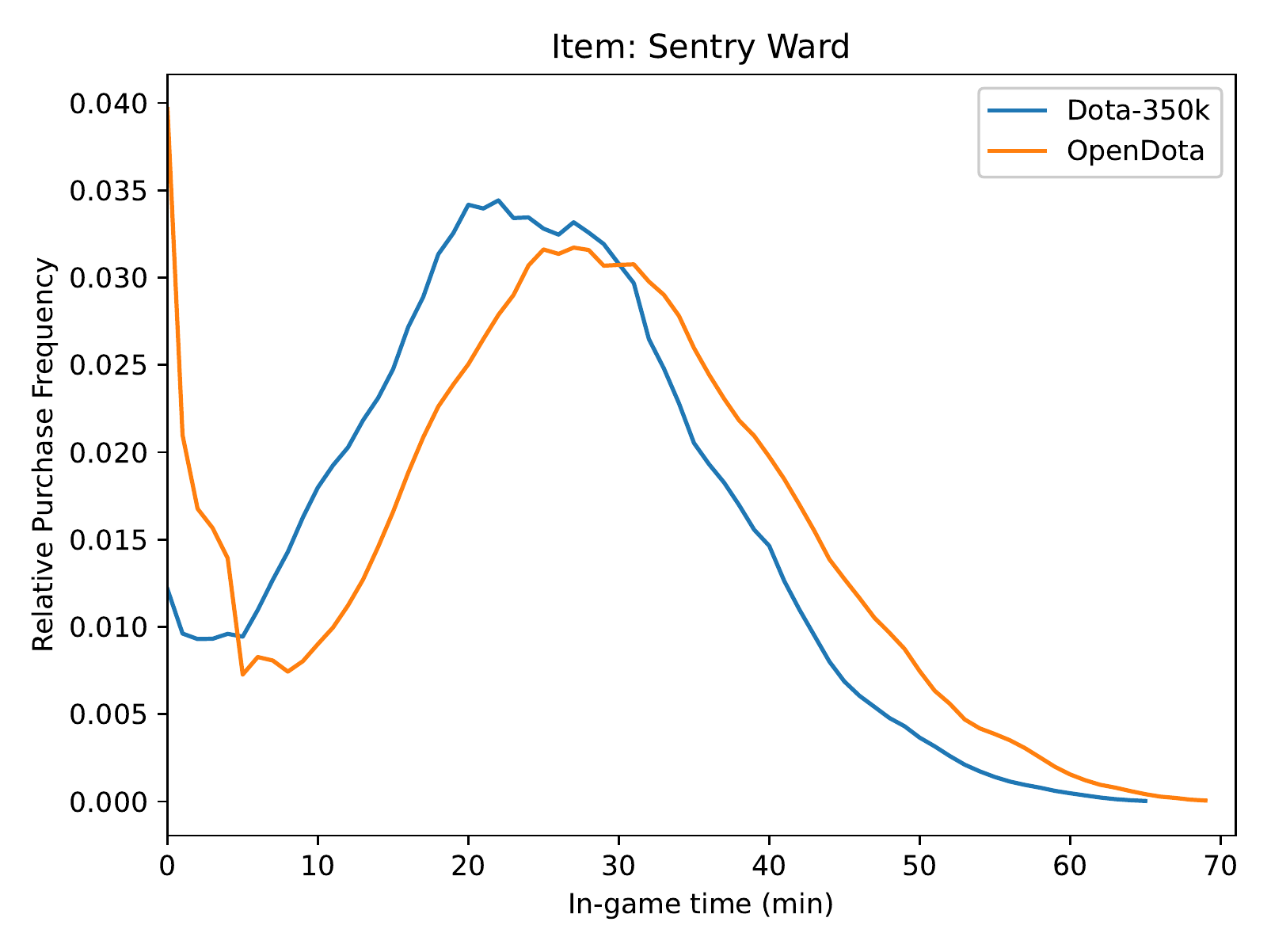}}
    \hfill
  \subfloat[Item Ethereal Blade\label{fig:blade_time_dist}]{%
        \includegraphics[scale=0.5]{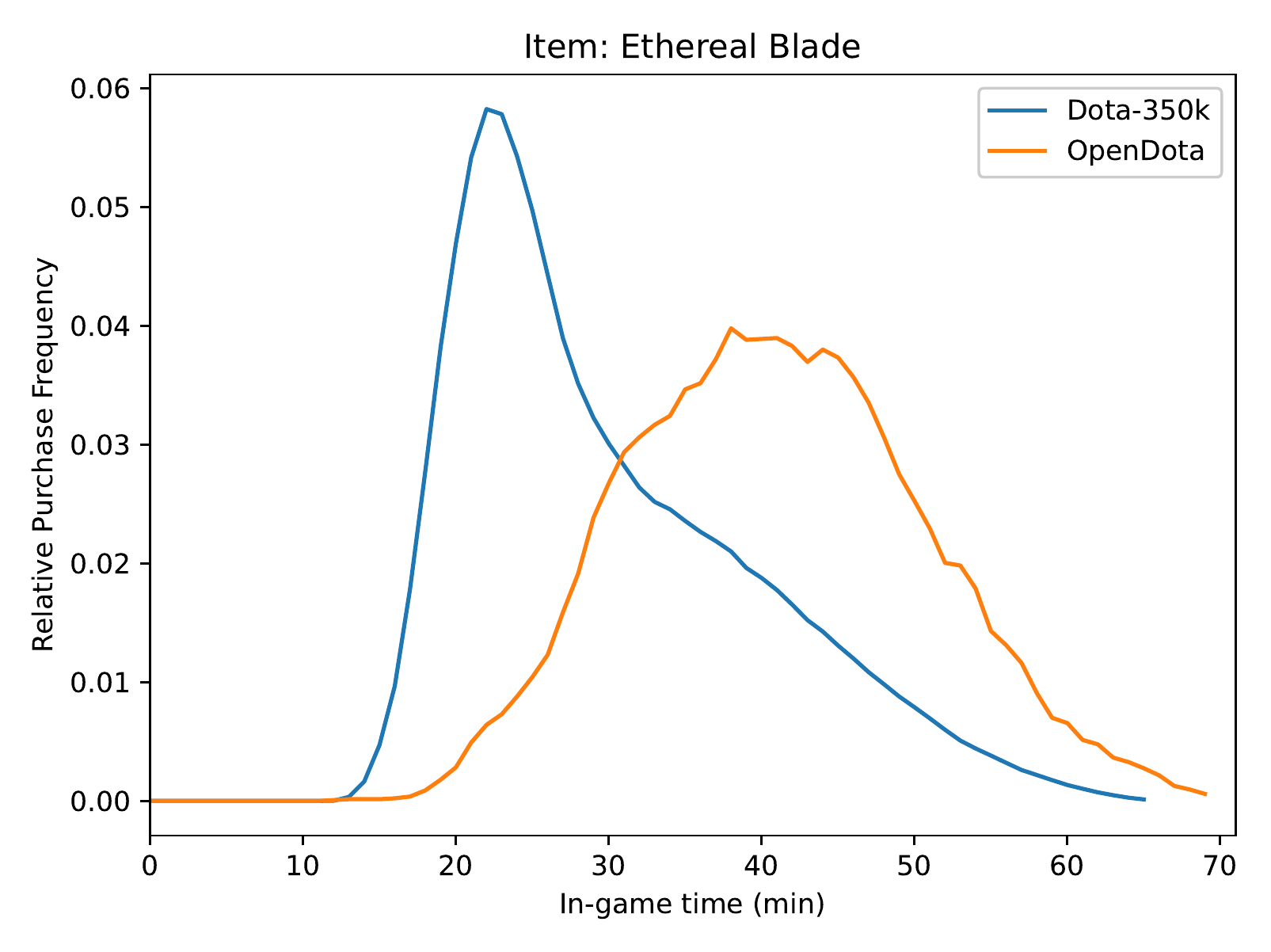}}
  \caption{The distribution of the purchase time of two items in in-game time. All purchase events are first binned into one minute intervals and normalized by the total number of purchases of that item in the respective dataset. Afterwards, a rolling average with a window size of 5 minutes is computed.}
  \label{fig:time_dist_figs} 
\end{figure}

\begin{figure}
    \centering
    \includegraphics[scale=0.5]{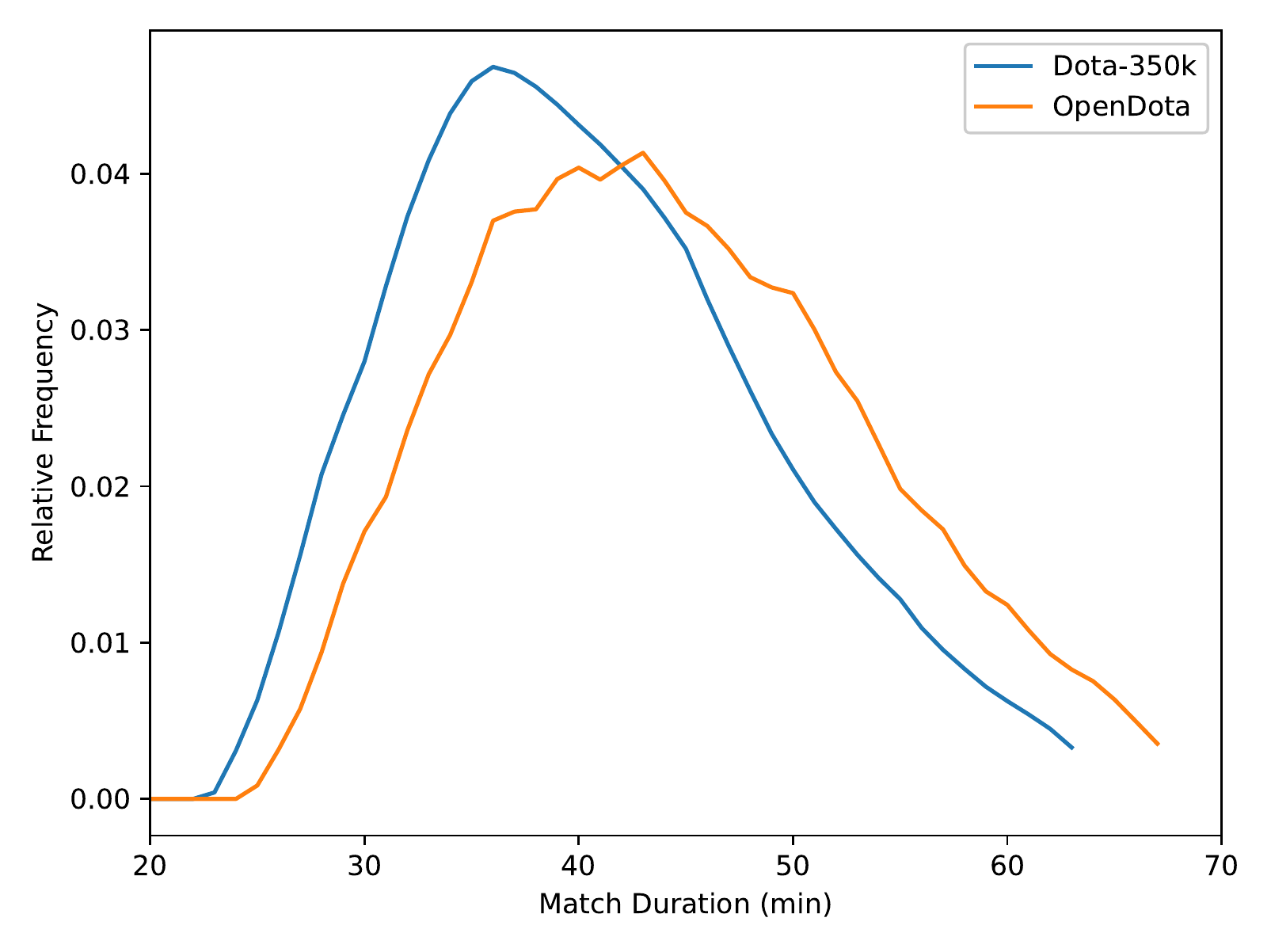}
    \caption{The distribution of the match durations in \sdota and \opendotas in in-game minutes. Match durations are first binned into one minute intervals and normalized by the total number of matches in the respective dataset. Afterwards, a rolling average with a window size of 5 minutes is computed.}
    \label{fig:match_duration}
\end{figure}

The \opendotas dataset available on \textit{Kaggle}\footnote{\url{https://www.kaggle.com/devinanzelmo/dota-2-matches}} was created from a dump of the \opendota\footnote{\url{https://www.opendota.com/}} database. The basic properties of the dataset can be found in \Cref{tab:opendota_statistics}.
\paragraph{Preprocessing and Split}
We apply the same preprocessing steps that were used for \sdota to \opendotas. Since the raw \opendota dataset consists of \textit{Ranked All Pick} matches only, no matches had to be removed due to this constraint. Abandonments, however, also affect \opendotas leading to the removal of $15.51\%$ of the originally contained matches. 
In contrast to \sdota, we use a standard 90\%/5\%/5\% split for \opendota. Due to the comparatively small size of \opendota, this is necessary to ensure that test and validation data closely resemble the hero and item distributions found in the training set. Although the split sizes differ, we still apply the same splitting methodology as described for \sdota.
\paragraph{Comparison to \sdota} 
While \sdota was compiled from recently gathered match data, \opendotas was built from matches collected in 2015 which equates to \dota patches 6.83c to 6.86c.
Since new heros and items were introduced into the game between the collection of the \opendotas dataset and \sdota, our dataset features more heros as well as a significantly larger number of items.
Due to the large difference in collection time, the in-game \textit{meta} also evolved significantly between the two datasets.
For example, comparing the purchase times of the item \textit{Sentry Ward} (see \Cref{fig:ward_time_dist}), reveals that it was most frequently bought at the start of the match in the \opendotas dataset, while being purchased the most approximately 30 minutes into the match in \sdota.
In contrast, \textit{Ethereal Blade} (see \Cref{fig:blade_time_dist}) is purchased earlier in matches on the more recent patch $7.25c$ then in those played in 2015.
The session characteristics presented in \Cref{tab:dataset_statistics} and \Cref{tab:opendota_statistics} show that the average number of items bought per session increases by approximately $23\%$ from $35$ to $43$.
The amount of gold earned per minute (GPM) is, however, similar in both datasets with approximately $415$ GPM in \opendotas and $405$ GPM in \sdota.
Since the average durations of a match only differ slightly between \sdota (39 min) and \opendota (43 min) (see \Cref{fig:match_duration}), the total amount of gold available to a player is roughly equal in both datasets.
Therefore, the difference in session length has to be caused by other factors such as cheaper items or increased purchase of consumables.
These \textit{meta} changes can also be observed via the pick frequency of individual heros (see \Cref{fig:item_and_hero_frequencies}).
The hero \textit{Windranger} who is the most popular hero in the \opendotas dataset only reaches rank 20 in \sdota while the most frequently picked hero in \sdota  -- \textit{Ogre Magi} -- is placed 40th in \opendota. The difference in pick frequency also implicitly influences the buy frequencies of individual items since some items are more frequently bought by specific heros. Furthermore, purchase frequencies are directly influenced by changes to the individual items. Therefore, changes in the in-game \textit{meta} influence the purchase behavior of the players, indicating that the performance of a recommender system might deteriorate when employed on more recent match data.



\section{Experiments}
\label{sec:experiments}
To judge the applicability of neural \sir models for item purchase recommendation in \dotatwo, we evaluate a set of well-known baselines and neural recommendation models on both datasets. Additionally, we examine the performance of a related approach recently proposed by \citeauthorwithcite{looi2019recommendation}.

\subsection{Models}
We implement all models using PyTorch Lightning\footnote{https://pytorch-lightning.readthedocs.io/en/latest/, Version: 1.4.0} as part of a comprehensive model library.
\subsubsection{Baselines}
We select two commonly used statistical baselines in order to establish a lower bound to judge the performance of the neural models. Furthermore, we evaluate a LR classifier proposed by \citeauthorwithcite{looi2019recommendation} in a similar setting. 
\paragraph{Most Popular}
\textit{\pop} (\pops) creates a static popularity ranking across all items by determining the buy frequency of each item in the dataset.
During evaluation the model then predicts the static popularity ranking independent of the previous interactions in the session.

\paragraph{Markov}
\textit{Markov} is a basic Markov Chain model that uses a transition matrix to predict the next item, based on the most recent interaction. The transition matrix is computed by aggregating all interactions for every item in the dataset and normalizing them to obtain transition probabilities for each pair of items. During inference, the probability distribution of the last item in the sequence is used to generate a ranked list of recommendations.

\paragraph{Logistic Regression}
Employing \textit{Logistic Regression} (LR) in the context of \dotatwos item recommendation was proposed by \citeauthorwithcite{looi2019recommendation}. For each buy event, a multi-hot coded feature vector is generated from the set of all previous purchases in the current match. Afterwards, a individual LR classifier is trained for each item using the aforementioned feature vectors. The predicted probabilities for each items are then used to generate a ranked list of recommendations during evaluation.

\subsubsection{Neural Models}
We evaluate the applicability of \sir models based on several well-established neural models. Additionally, we employ a MLP to examine the influence of model complexity on recommendation performance.
\paragraph{Multilayer-Perceptron}
We implement a basic \textit{Multilayer-Perceptron} (MLP) that receives the same multi-hot coded feature-vector based on the previous purchases in the sessions, as the LR  model. 
The MLP model uses \textit{\relu} activation in its hidden layers. 
It is trained to predict the next item using a standard cross-entropy loss.
Compared to the LR model, the MLP model introduces non-linearity and capacity scaling to the non-sequential input setting and helps to judge the improvement gained from treating the input as a sequence.

\paragraph{GRU}
\label{par:gru}
Similar to \citeauthorwithcite{tan2016improved} we use a Recurrent Neural Network with \gru memory cells to encode the sequence of purchased items.
The final hidden state in the sequence is projected into the item space and the result is used to predict the next item. 
The model is trained using a cross-entropy loss. 

\paragraph{NARM}
The Neural Attentive Recommendation Model (\narm) published by \citeauthorwithcite{li2017neural} uses separate global and local encoder layers to encode the sequence of purchased items. 
Both encoder layers use a RNN with \gru cells to encode the sequence of items. 
The final hidden state of the global encoder (\textit{global encoding}) is then used to compute attention scores for the sequence of hidden states from the local encoder which are used to compute a \textit{local encoding} through a weighted sum. 
Both encodings are concatenated and the resulting vector is projected into the item space using a bilinear decoding scheme.
The model is then trained using the cross-entropy loss.

\paragraph{SASRec}
Self-Attentive Sequential Recommendation (SASRec) \cite{kang2018selfattentive} is based on the transformer encoder module introduced by \citeauthorwithcite{vaswani2017attention}.
Several encoder modules are stacked to form a deep transformer network.
The model simultaneously makes a prediction for every step in the sequence and is
is trained using the BPR loss function~\cite{rendle2009bayesian}.

\paragraph{BERT4Rec}
BERT4Rec \cite{sun2019bert4rec} adapts the BERT model \cite{Devlin2019BERTPO} from language modeling to the task of sequential recommendation.
The model uses a bidirectional transformer encoder and is trained by predicting randomly masked positions in the sequence (Cloze task \cite{taylor1953cloze}).
\\
\indent While \pops, \sasrec and \bertforrec are trained by recommending an item for each position in the session simultaneously, the remaining models are trained by only predicting the next item. To increase the number of available training instances and to cover all positions in each session, we augment the training samples to include all sub-sequences of each session \cite{tan2016improved}.

\subsection{Evaluation Metrics}
We use \textit{Recall@$k$} and \textit{NDCG@$k$}, two common evaluation metrics to evaluate the model performance \cite{sun2019bert4rec, kang2018selfattentive}.
Recall@$k$ measures whether the target item is part of the top-$k$ items returned by the model, disregarding the actual rank of the target item. 
In contrast, NDCG@$k$ uses the actual target item rank to award scores based on the rank in the result list, with lower ranks earning lower scores.
Since \dota is a fast paced game, where quick decision making is important, it makes sense to present the player only with a small set of possible choices.
Therefore, we evaluate all trained models with $k=1,3$ for both metrics.

\subsection{Hyper Parameter Study}
\label{sec:hyper_parameter_study}
To tune the hyper-parameters of the neural models, we perform a random search on the hyper-parameter space defined in~\Cref{tab:hyperparameters}.
On both datasets, we train $30$ instances of each model and select the best model according to the metric \textit{Recall@3}.
In addition to the parameters that are sampled uniformly at random from the ranges listed in ~\Cref{tab:hyperparameters}, we use early stopping with the patience set to $p=3$ and the maximum number of epochs set to $25$.
Since the learning rate did not have a significant impact in initial experiments, we set it to $lr=10^{-3}$ for all models and the number of warm up steps to $10000$ for BERT4Rec.
In order to train transformer based models efficiently, the maximum sequence length needs to be restricted to a sufficiently small value.
We determine the maximum sequence length for each dataset as the length that fully includes $\num{99.5}\%$ off all sessions, (i.e., $l_{s, max} = 63$ for \opendota and $l_{s, max}=90$ for \sdota), and truncate the remaining $\num{0.5}\%$ sessions to these lengths by removing the start of the session.
To allow a fair comparison, we restrict all neural models to the same maximum input sequence length.
In the rest of the paper we only report results for the best model settings discovered during the search (see \Cref{tab:hyperparameters}).

\begin{table}
    \caption{Ranges for the hyper parameters used during the random search for each neural model and the settings for the best model found by the search for each dataset.(*) Note that the actual hidden size is computed as $h = \#heads \times hidden size$ due to model constraints.}
    \centering
    \begin{tabular}{llcccc}
        \toprule
        & parameter  & range & step size & \opendota & \sdota \\
        \midrule
        \parbox[t]{2mm}{\multirow{2}{*}{\rotatebox[origin=c]{90}{MLP}}} & hidden size  & $[16,256]$ & 16 & 256 & 256 \\
        & \#layers  & $[1, 3]$ & 1 & 2 & 3\\
        \midrule
        \parbox[t]{2mm}{\multirow{5}{*}{\rotatebox[origin=c]{90}{BERT4Rec}}} & \#heads  & $[1,8]$ & 1 & 8 & 7 \\
        & \#layers  & $[1, 6]$ & 1 & 4 & 5 \\
        & hidden size* & $[8, 32]$ & 1 & 32 & 17\\
        & dropout & $[0.1, 0.5]$ & 0.05 & 0.1 & 0.1\\
        & non-linearity  & $[\relu, \tanh]$ & 1 & \relu & \relu\\
        \midrule
        \parbox[t]{2mm}{\multirow{5}{*}{\rotatebox[origin=c]{90}{SASRec}}} & \#heads  & $[1,8]$ & 1 & 5 & 7\\
        & \#layers  & $[1, 6]$ & 1 & 6 & 4 \\
        & hidden size* & $[8, 32]$ & 1 & 30 & 13 \\
        & dropout & $[0.1, 0.5]$ & 0.05 & 0.2 & 0.1 \\
        & non-linearity  & $[\relu, \tanh]$ & 1 & $\tanh$ & $\tanh$\\
        \midrule
        \parbox[t]{2mm}{\multirow{4}{*}{\rotatebox[origin=c]{90}{GRU}}} & emb. size  & $[16, 64]$ & 16 & 32 & 64 \\
        & cell size & $[16, 256]$ & 16 & 224 & 128 \\
        & \#layers  & $[1, 2]$ & 1 & 2 & 2 \\
        & dropout & $[0.1, 0.5]$ & 0.05 & 0.1 & 0.1\\
        \midrule
        \parbox[t]{2mm}{\multirow{5}{*}{\rotatebox[origin=c]{90}{NARM}}} & emb. size  & $[16, 64]$ & 16 & 64 & 32 \\
        & enc. size & $[16, 256]$ & 16 & 176 & 80 \\
        & \#enc. layers  & $[1, 2]$ & 1 & 1 & 1 \\
        & ctx. dropout & $[0.1, 0.5]$ & 0.05 & 0.15 & 0.2 \\
        & emb. dropout & $[0.1, 0.5]$ & 0.05 & 0.1 & 0.15 \\
        \bottomrule
    \end{tabular}
    \label{tab:hyperparameters}
\end{table}

\subsection{Results and Discussion}
\label{sec:results}

\Cref{tab:results} shows the model performance of the best hyper-parameter settings for both datasets using the metrics Rec@$k$ and NDCG@$k$ for $k=1,3$.
All models perform slightly worse on \sdota when compared to \opendota.
This is likely due to the fact, that \sdota has an approximately $69\%$ bigger item space than \opendota.
Additionally, in \sdota, the top three most frequently bought items cover only ~$28.75\%$ of all purchases compared to ~$33.75\%$ in \opendota, which directly affects the \pops baseline and is likely to also negatively impact the other models.
The relative performance ranking of the models does not change between both datasets, allowing us to transfer all statements about relative performance from one dataset to the other.
Consequently, we will only include results from \sdota in the following discussion.

\paragraph{Baselines}
\pops predicts a static ranking according to the overall item purchase frequency and thus does not consider any information about the current session, \ie already purchased items.
The results show that this approach yields the weakest performance with Rec@3 = $0.2942$.
Including the previously purchased item into the \markov model yields a performance improvement to Rec@3 = $0.5198$.
The \lrs model improves the performance further to Rec@3 = $0.5790$ by using the information about all items bought previously in the session, although it discards the order of purchase events and can't model buying an item multiple times (which is not uncommon do to \textit{consumables}).
These results show that latent dependencies between items bought during a match exist and can be used to improve the prediction quality of models.

\paragraph{Neural Models}
The neural models can be divided into three categories: feed-forward, RNN-based and transformer-based.
Our results show that the RNN-based models, \ie \gru (Rec@3 = $0.7363$) and \narm (Rec@3 = $0.6853$), clearly outperform the transformer models on both datasets.
Although \bertforrec and \sasrec have been shown to yield better performance on datasets from other domains\cite{kang2018selfattentive,sun2019bert4rec}, this is not the case for purchases in \dotatwo matches.
Similarly to the LR classifier, the MLP model uses the set of previously purchased items as input and is able to improve upon the results of all baselines including the LR classifier achieving a score of Rec@3 = $0.6608$. It is, however, outperformed by both RNN-based models. Since the MLP model hyper parameters have been tuned, this indicates, that the performance gain of models that consider the sequence of purchases instead of treating them as a set does not simply stem from increased model complexity but from the explicit inclusion of sequential information.
Both \sasrec and \bertforrec are based on a transformer encoder architecture, with the difference that \bertforrec uses a bidirectional encoder. It is, therefore, surprising that \sasrec (Rec@3 = $0.4118$) performs even worse then the \markov model which only bases the prediction on the last interaction.
Besides model architecture, a key difference between \bertforrec and \sasrec is the type of loss function used during training.
While \sasrec uses the next-item prediction task with BPR\cite{rendle2009bayesian}, a type of negative sampling loss, \bertforrec is based on the Cloze task\cite{taylor1953cloze} and uses a cross-entropy loss.
We hypothesize, that the comparatively small item space may negatively impact the performance of the negative sampling loss during training and thus prevent the model from achieving a competitive prediction performance.
It is noteworthy that \bertforrec only consistently outperforms the MLP model for Rec@1 and NDCG@1 while the results for Rec@3 and NDCG@3 are mixed.

\begin{table}
    \centering
    \caption{Results for the \opendotas and \sdota dataset in terms of Recall@$k$ and NDCG@$k$ with $k={1,3}$ for the baseline models, \pops, Markov and LR, as well as for the best parameterization of the considered neural recommendation models: MLP, GRU, SASRec, Bert4Rec and NARM.}
  \begin{subtable}{\linewidth}
  \centering
        \begin{tabular}{lcccccc}
            \toprule
                                     &   Rec@1 / NDCG@1     &   Rec@3       & NDCG@3  \\
            \midrule
            POP                      & \num{0.1195} & \num{0.2942}  & \num{0.2194}  \\
            Markov                   & \num{0.3001} & \num{0.5198}  & \num{0.4275}  \\
            LR~\cite{looi2019recommendation} & \num{0.3164} & \num{0.5790} & \num{0.4678}  \\
            \midrule
            MLP                      & \num{0.3858} & \num{0.6608}  & \num{0.5447}  \\
            GRU                      & \textbf{\num{0.4825}} & \textbf{\num{0.7363}}  & \textbf{\num{0.6307}}  \\
            NARM                     & \num{0.4348} & \num{0.6853}  & \num{0.5808}  \\
            SASRec                   & \num{0.1522} & \num{0.4118}  & \num{0.2996}  \\
            Bert4Rec                 & \num{0.4117} & \num{0.6480}  & \num{0.5492}  \\
            \bottomrule
        \end{tabular}
        \caption{\sdota}
        \label{tab:results-dota}
    \end{subtable}
    \begin{subtable}{\linewidth}
        \centering
        \begin{tabular}{lccc}
            \toprule
                                     &   Rec@1 / NDCG@1      &   Rec@3   & NDCG@3  \\
            \midrule
            POP                      & \num{0.2354} & \num{0.3433}  & \num{0.2983}  \\
            Markov                   & \num{0.3068} & \num{0.5439}  & \num{0.4440}  \\
            LR~\cite{looi2019recommendation} & \num{0.3598} & \num{0.6513} & \num{0.5299} \\
            \midrule
            MLP                      & \num{0.4222} & \num{0.7212} & \num{0.5968} \\
            GRU                      & \textbf{\num{0.5074}} & \textbf{\num{0.7833}} & \textbf{\num{0.6696}} \\
            NARM                     & \num{0.4970} & \num{0.7736} & \num{0.6596}  \\
            SASRec                   & \num{0.2442} & \num{0.4466} & \num{0.3589}  \\
            Bert4Rec                 & \num{0.4359} & \num{0.7054} & \num{0.5935}  \\
            \bottomrule
    \end{tabular}
    \caption{\opendotas}
    \label{tab:results-opendota}
    \end{subtable}
    \label{tab:results}
    \vspace{-2em}
\end{table}

\section{Conclusion}
\label{sec:conclusion}
In this work, we introduced a novel, large-scale dataset of purchase sessions compiled from interactions of \dotatwos players with the in-game store. 
In comparison to the already existing \opendotas dataset, \sdota contains approximately nine times as many matches recorded on a recent patch version.
Therefore, it also features more heros and a larger item space than \opendotas.
In order to examine the applicability of sequential item recommendation for in-game purchase advise, we evaluated the performance of statistical baselines, a related set-based approach using LR and several neural recommendation models. 
While the neural models generally outperformed the baselines by a large margin, the performance of \sasrec surprisingly was subpar in comparison to related domains \cite{kang2018selfattentive,sun2019bert4rec}.
The RNN-based models \gru and \narm surpassed the transfromer-based architectures with \gru achieving the highest scores throughout both datasets.
Also note that all sequence-based neural models -- except \sasrec\space-- generally outperform the set-based MLP model and the LR model proposed by \citeauthorwithcite{looi2019recommendation}, indicating that the purchase order is a relevant feature that improves model performance.

Even though we have shown \sir systems to be suited for item recommendation in \dotatwos, they do not leverage all data present in a \dotatwos match. 
Since the replay file of a match includes all information necessary to completely replay the match, a variety of contextual information is available. 
While the purchase history forms an important basis for a purchase decision, factors such as the draft of the allied and enemy team, the in-game time as well as the purchases of all other players play an important role in correct itemization.
In future work, we will focus on including different types of contextual information extracted from \dotatwo matches into recommendation models.

\renewcommand*{\bibfont}{\footnotesize}
\printbibliography

\end{document}